\documentclass[times,twocolumn,final]{elsarticle}

\usepackage{medima}
\usepackage{framed,multirow}

\usepackage{amssymb}
\usepackage{amsmath}
\usepackage{latexsym}
\usepackage{booktabs}
\usepackage{multirow}
\usepackage{siunitx}
\usepackage{bbm}

\usepackage{url}
\usepackage{xcolor}

\usepackage{hyperref}

\definecolor{mygreen}{RGB}{0,127,0}
\definecolor{myred}{RGB}{255,0,0}
\definecolor{myblue}{RGB}{0,0,255}
\definecolor{rev_changes}{RGB}{0,0,0} 

\journal{Medical Image Analysis}

\begin{document}

\verso{Alexander Bigalke \textit{et~al.}}

\begin{frontmatter}

\title{Anatomy-guided domain adaptation for 3D in-bed human pose estimation}%

\author[1]{Alexander \snm{Bigalke}\corref{cor1}}
\cortext[cor1]{Corresponding author: 
  Tel.: +49 451 3101 5619;}
\ead{alexander.bigalke@uni-luebeck.de}
\author[2]{Lasse \snm{Hansen}}
\author[3]{Jasper \snm{Diesel}}
\author[4]{Carlotta \snm{Hennigs}}
\author[4]{Philipp \snm{Rostalski}}
\author[1]{Mattias P. \snm{Heinrich}}

\address[1]{Institute of Medical Informatics, University of L\"ubeck, Ratzeburger Allee 160, 23538 L\"ubeck, Germany}
\address[2]{EchoScout GmbH, Maria-Goeppert-Str.~3, 23562 L\"ubeck, Germany}
\address[3]{Dr\"agerwerk AG \& Co.~KGaA, Moislinger Allee 53-55, 23558 L\"ubeck, Germany}
\address[4]{Institute for Electrical Engineering in Medicine, University of L\"ubeck, Moislinger Allee 53-55, 23558 L\"ubeck, Germany}

\received{-}
\finalform{-}
\accepted{-}
\availableonline{-}
\communicated{-}

\begin{abstract}
3D human pose estimation is a key component of clinical monitoring systems.
The clinical applicability of deep pose estimation models, however, is limited by their poor generalization under domain shifts along with their need for sufficient labeled training data.
As a remedy, we present a novel domain adaptation method, adapting a model from a labeled source to a shifted unlabeled target domain.
Our method comprises two complementary adaptation strategies based on prior knowledge about human anatomy.
First, we guide the learning process in the target domain by constraining predictions to the space of anatomically plausible poses.
To this end, we embed the prior knowledge into an anatomical loss function that penalizes asymmetric limb lengths, implausible bone lengths, and implausible joint angles.
Second, we propose to filter pseudo labels for self-training according to their anatomical plausibility and incorporate the concept into the Mean Teacher paradigm.
We unify both strategies in a point cloud-based framework applicable to unsupervised and source-free domain adaptation.
Evaluation is performed for in-bed pose estimation under two adaptation scenarios, using the public SLP dataset and a newly created dataset.
Our method consistently outperforms various state-of-the-art domain adaptation methods, surpasses the baseline model by 31\%/66\%, and reduces the domain gap by 65\%/82\%.
Source code is available at \url{https://github.com/multimodallearning/da-3dhpe-anatomy}.
\end{abstract}


\end{frontmatter}


\section{Introduction}
3D human pose estimation is a fundamental problem in computer vision and the basis for various higher-level tasks, such as posture recognition \citep{liu20203d} and action recognition \citep{song2021human}.
\begin{figure*}[t]
\begin{center}
  \includegraphics[width=\linewidth]{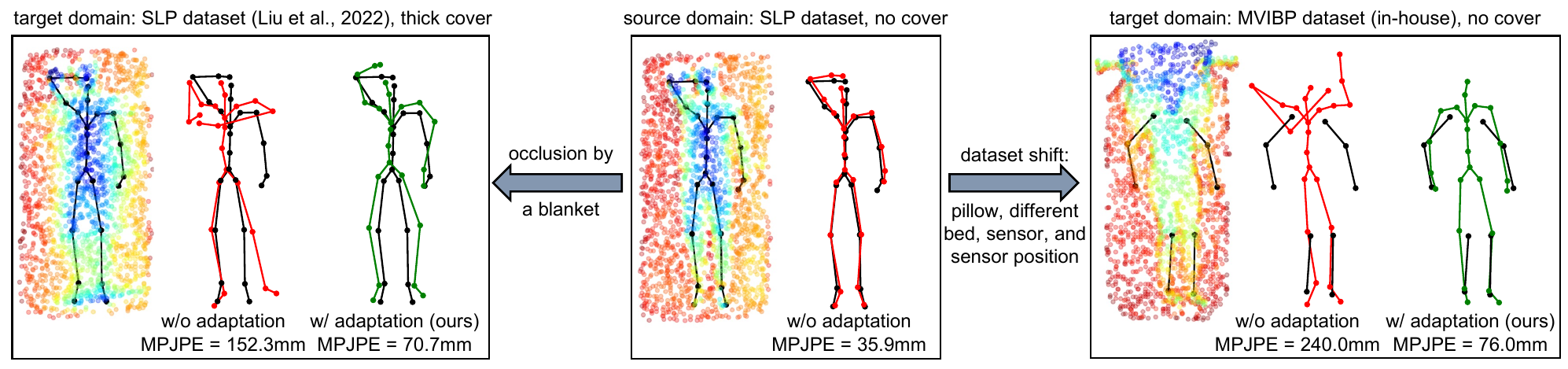}
\end{center}
  \caption{Visualization of two domain shifts for point cloud-based in-bed pose estimation and their impact on model performance.
  We show input point clouds from the source domain and two different target domains (colors encode the depth in z-direction) alongside the ground truth poses in black, the predictions by a source-trained baseline model in red, and the predictions by our adaptation method in green.
  While the in-domain prediction of the baseline model is close to perfect, the predictions on the shifted domains are anatomically implausible and highly inaccurate (in terms of the mean per joint position error MPJPE).
  Adaptation with our anatomy-guided method substantially improves the accuracy and plausibility of the pose estimates.}
\label{Fig:motivation}
\end{figure*}
These tasks, in turn, open up a wide range of applications in the field of human-computer interaction, which are in high demand in the automotive or gaming sectors, for instance \citep{chen2020monocular}.
The healthcare sector can also benefit from automatic pose estimation as pose-based assistance and monitoring systems promise to relieve clinical staff and improve patient safety and care.
On the one hand, tracking the 3D joint positions of clinicians enables automated documentation, analysis, and optimization of clinical workflows \citep{mascagni2021or,rodrigues2022multi}.
On the other hand, in-bed pose estimation, the application focus of this work, offers great potential for automatic patient monitoring:
A pose-driven monitoring system could analyze movements \citep{chen2018patient}, detect potentially critical events \citep{jahne2019inbed}, diagnose pathological movement patterns \citep{cunha2016neurokinect}, and prevent pressure ulcers \citep{ostadabbas2012resource}.

In recent years, deep learning has substantially advanced the state of the art in general and clinical human pose estimation \citep{chen2020monocular}, making the deployment of the above systems more tangible.
Nonetheless, several challenges remain, particularly in the clinical setting.
First, data privacy and highly variable lighting conditions, including complete darkness, preclude the use of standard color images.
As a remedy, we advocate the use of 3D point cloud data.
Point clouds are not only anonymity-preserving \citep{silas2015video} and insensitive to lighting conditions but also inherently preserve the 3D structure of the scene, making them a natural modality for 3D pose estimation.
Second, the performance of deep learning-based methods strongly depends on access to large-scale labeled datasets \citep{ionescu2013human3}.
The annotation of 3D poses, however, is generally laborious and even more involved in clinical settings:
Data access is often restricted, and accurate annotations under severe occlusions, e.g., caused by blankets in the case of patient monitoring, are only feasible under the controlled conditions of a lab study \citep{liu2022simultaneously}.
Therefore, it is crucial to take full advantage of existing datasets \citep{liu2022simultaneously,srivastav2018mvor} as a training resource across diverse target domains.
However, this is hampered by the poor generalization of deep models under domain shifts, resulting in severe performance drops when deploying a model in a shifted domain \citep{wang2021human}.
In the clinical setting, such shifts can be due to varying room setups/environments in different hospitals/countries or changing visibility conditions (no blanket, blanket), as visualized in Fig.~\ref{Fig:motivation}.
While supervised fine-tuning on shifted data could alleviate the problem, it is often no viable solution given the high annotation costs.
Instead, it is desirable to adapt a model from a labeled source to an unlabeled target domain in an unsupervised fashion.
This can be realized by domain adaptation (DA) \citep{wang2018deep}, the methodological focus of this work.

Classical unsupervised domain adaptation (UDA) methods approach the problem by jointly accessing data from both domains.
Given the importance of data protection in the medical sector, however, this cannot always be guaranteed.
Instead, it is a realistic scenario that the provider of a pose estimation model and its end-user are not able or willing to exchange their data.
Consequently, the end-user needs to adapt the provided pretrained source model to the target domain without accessing the source data, denoted as source-free domain adaptation (SFDA) \citep{kundu2020universal}.
With this in mind, it is desirable to have a universal DA method applicable to both UDA and SFDA as needed.

A popular branch of DA methods couples the supervised learning on labeled source data with the alignment of the distributions of source and target features, realized by discrepancy minimization \citep{tzeng2014deep} or adversarial learning \citep{ganin2015unsupervised,tzeng2017adversarial}.
The learned target features, however, are not explicitly optimized for the actual task, and domain invariance does not guarantee task relevance.
This problem can be addressed by performing the adaptation in the output space of the target domain, implemented by adversarial optimization \citep{tsai2018learning,yang20183d} and direct supervision with pseudo labels \citep{mu2020learning,yang2021st3d} in prior work.
However, adversarial optimization is complex and unstable, and pseudo labels are noisy and can thus misguide the learning process.
As an additional downside, adversarial methods are not applicable to SFDA since they require simultaneous access to both domains.

We propose to overcome these problems by guiding the adaptation process with the aid of prior knowledge about human anatomy.
Such prior knowledge contains valuable information about the expected pose distribution in the output space, which can be vastly restricted by excluding anatomically implausible poses that cannot be taken by a human.
Notably, the prior knowledge is domain-independent and thus invariant under the domain shifts discussed above.
We propose two different strategies to exploit this knowledge (see Fig.~\ref{Fig:method_overview} for an overview).
First, we directly supervise predictions in the target domain by explicitly constraining them to the space of anatomically plausible poses.
To this end, we derive three anatomical loss functions that penalize predictions with asymmetric limb lengths, implausible bone lengths, and implausible joint angles.
Second, we filter noisy pseudo labels for self-training according to their anatomical plausibility, measured with our anatomical loss functions.
Concretely, we incorporate this technique in the Mean Teacher paradigm \citep{french2017self,tarvainen2017mean}, where pseudo labels from the teacher are only used for supervision if they are more plausible than the current prediction of the learning student model.
We unify these two strategies in a point cloud-based framework.
It performs output adaptation without intricate adversarial optimization and mitigates noisy supervision through anatomical guidance.
Moreover, it does not require simultaneous access to source and target domain and is thus applicable to both UDA and SFDA.

In summary, the main contributions of this work are:
\begin{enumerate}
\item We introduce an anatomy-guided domain adaptation method for point cloud-based 3D human pose estimation, including two complementary adaptation strategies based on prior anatomical knowledge.
\item We derive an anatomical loss function that constrains pose predictions in the target domain to the space of plausible poses by penalizing asymmetric limb lengths, implausible bone lengths, and implausible joint angles.
\item We propose to filter pseudo labels based on their anatomical plausibility and incorporate the concept into the Mean Teacher paradigm. 
\item We demonstrate the efficacy of our method in the context of in-bed pose estimation for both UDA and SFDA under two different scenarios: the adaptation between the different environments of two datasets---the public SLP dataset \citep{liu2019seeing,liu2022simultaneously} and a newly created dataset---and from uncovered to covered patients. Under all settings, our method is superior to a comprehensive set of state-of-the-art domain adaptation methods, which we adapted to the given problem.
\end{enumerate}

A preliminary conference version of this work appeared at MIDL 2022 \citep{bigalke2021domain}.
In this journal version, we extend this work as follows:
1) We substantially extend the discussion of related works.
2) We give a more detailed description of the method and derive the anatomical loss function from a constrained optimization problem.
3) Extending the method, we use the anatomical loss not only for direct supervision but propose to use it as a criterion for filtering pseudo labels.
4)~We formalize anatomy-constrained optimization and anatomy-guided filtering of pseudo labels in a unified framework applicable to UDA and SFDA.
5) We perform extensive additional experiments, demonstrating the efficacy of our method under a second adaptation scenario (using our recently captured dataset) and in the challenging SFDA setting.

\section{Related work}
\subsection{Human pose estimation}
2D and 3D human pose estimation from regular 2D grid data is a widely studied problem, with most works focusing on RGB \citep{sun2019deep,xiao2018simple} and depth images \citep{haque2016towards,moon2018v2v} as the input modalities.
Since our work treats point cloud-based pose estimation (see Sec.~\ref{Sec:pcd_pose_estimation}), we refer the reader to \citet{chen2020monocular} for a comprehensive survey of grid-based methods and summarize works with clinical applications.
The first line of such works addresses pose estimation of clinical staff in the operating room.
While early methods rely on multi-view RGB \citep{belagiannis2016parsing} and RGB-depth \citep{kadkhodamohammadi2017multi} images, more recent methods exploit multi-view \citep{hansen2019fusing} and low-resolution \citep{srivastav2019human} depth images to prevent privacy concerns by clinicians and patients.
Another stream of methods treats in-bed patient pose estimation.
Besides compliance with data protection, the primary challenge in this task consists of severe occlusions by blankets.
Multiple works aim to see under the blanket with the help of suitable sensors.
\citet{liu2019seeing} estimate 2D poses from thermal images, and \citet{casas2019patient,davoodnia2021bed} use pressure maps to estimate 3D and 2D poses, respectively.
Alternatively, several methods learn to predict the pose and shape parameters of a human mesh model \citep{loper2015smpl} under blanket occlusions by fusing multiple modalities, including thermal, pressure, depth, and RGB images \citep{karanam2020towards,yang2020robust,yin2020multimodal}.
However, all the above methods require ground truth annotations under the blanket, which are difficult to obtain in a real-world application.
As a remedy, \citet{achilles2016patient,clever2020bodies,clever2022bodypressure} train their models on synthetic depth or pressure maps of covered patients, and \citet{afham2022towards,chi2022multi} perform domain adaptation from labeled uncovered to unlabeled covered subjects based on thermal images (see Sec.~\ref{Sec:DA_pose_estimation}).

\subsection{Point cloud-based pose estimation}
\label{Sec:pcd_pose_estimation}
Compared to all the above modalities, point clouds stand out by inherently preserving the 3D structure of the scene.
Their unstructured nature, however, prevents the use of standard convolutions, complicating the processing with deep neural networks.
The pioneering PointNet \citep{qi2017pointnet} addressed the issue by extracting point-wise spatial representations, which are aggregated by max-pooling.
To capture local geometric structures, various follow-up works proposed hierarchical grouping \citep{qi2017pointnet++} and generic convolutions \citep{li2018pointcnn,liu2019relation,wang2019dynamic,wu2019pointconv,xu2021paconv} applicable to unstructured data.

Prior works on point cloud-based keypoint estimation primarily focus on hand pose estimation.
The Hand PointNet \citep{ge2018hand} employs the PointNet++ \citep{qi2017pointnet++} architecture for direct regression of the joint coordinates, followed by a refinement network for the fingertips.
In another work, \citet{ge2018point} extend PointNet++ to a stacked hourglass architecture \citep{newell2016stacked} and estimate joint coordinates by combined regression of heatmaps and offset vectors.
\citet{li2019point} regress separate pose estimates from the representations of each input point, which are aggregated in a final estimate.
\citet{hermes2022support} reduce the complexity of the regression problem by predicting joint coordinates as the weighted sum over the input points, complemented by a set of support points.
In our work, we employ the Dynamic Graph CNN (DGCNN) by \citet{wang2019dynamic} as the backbone architecture and formulate human pose regression similar to \citet{hermes2022support}.

\subsection{Domain Adaptation}
Classical UDA assumes joint access to a labeled source and a shifted unlabeled target domain.
We broadly classify UDA methods according to the level where the adaptation is performed: the input level, the feature level, and the output level.
The idea of input-level adaptation \citep{hoffman2018cycada,li2019bidirectional,murez2018image} is to align the image styles or pixel-level distributions of source and target data through image-to-image translation modules like CycleGAN \citep{zhu2017unpaired} or CUT \citep{park2020contrastive}.
By contrast, feature-level adaptation aims at aligning intermediate feature distributions from the source and target domain.
This was realized by minimizing explicit distance measures between both distributions \citep{rozantsev2018beyond,sun2016return,tzeng2014deep}, by adversarial learning with a domain discriminator \citep{ganin2015unsupervised,tzeng2017adversarial,saito2019strong}, and by simultaneously learning an auxiliary self-supervised task in both domains \citep{bousmalis2016domain,ghifary2016deep,sun2019unsupervised}.
Finally, \citet{luo2019taking,tsai2018learning} proposed to align source and target distributions in the output space by training the entire task network in an adversarial manner against a discriminator.

An alternative technique for output-level adaptation is self-training with pseudo labels \citep{zou2018unsupervised}.
The basic idea is to alternately generate pseudo labels on unlabeled target data with the current model and to re-train the model using these labels.
A specific form of self-training is the Mean Teacher paradigm \citep{tarvainen2017mean}, where pseudo labels are continuously generated by a teacher model, whose weights are given as the exponential moving average of the weights of the learning student network.
Initially introduced for semi-supervised classification, the concept was transferred to domain adaptation by \citet{french2017self} and subsequently adapted to diverse tasks, including object detection \citep{cai2019exploring,deng2021unbiased}, medical image segmentation \citep{li2020dual,perone2019unsupervised}, and medical registration \citep{bigalke2022adapting}.
However, pseudo labels are typically noisy, which can hamper the adaptation process.
Therefore, multiple works guide the supervision with pseudo labels through uncertainty estimates, computed by Monte Carlo Dropout \citep{wang2020double,wang2021tripled,yu2019uncertainty}, as the predictive variance under input perturbations \citep{zhou2022uncertainty} and among different network heads \citep{zheng2020cartilage,zheng2021rectifying}, and as the reconstruction error of a denoising autoencoder \citep{adiga2022leveraging}. 

Unlike UDA, SFDA aims to adapt a pre-trained source model to the target domain without accessing source data. Thus, the explicit alignment of both domains is no longer feasible.
To overcome this problem, \citet{kurmi2021domain,liu2021source} generate synthetic source data by exploiting the pre-trained source model.
In a different approach, the source model is directly adapted to the target domain by entropy minimization \citep{wang2020tent}, entropy minimization guided by shape priors \citep{bateson2020source}, and information maximization \citep{liang2020we}.
Similar to UDA, self-training with reliable \citep{kundu2021generalize} or denoised \citep{chen2021source} pseudo labels and the Mean Teacher \citep{hegde2021uncertainty,wang2022continual} were also deployed in source-free settings.
Another line of works achieved SFDA by progressively adapting the statistics of the BatchNorm layers to the target domain \citep{klingner2022unsupervised,liu2021adapting,zhang2020generalizable}.

\subsection{Point cloud-based domain adaptation}
The vast majority of point cloud-based DA methods perform feature-level adaptation through self-supervision and mainly differ by the pretext tasks.
The proposed tasks include the reconstruction of a deformed point cloud \citep{achituve2021self}, solving 3D puzzles \citep{alliegro2021joint}, and learning the implicit function that represents the underlying shape model \citep{shen2022domain}.
Some works suggested multi-level self-supervised learning at global and local scales \citep{fan2022self,zou2021geometry}: global tasks are scale and rotation prediction, while local tasks consist in the reconstruction of local areas and the localization of local distortions.
Besides self-supervised DA, \citet{qin2019pointdan} proposed multi-level alignment of local and global features, and \citet{cardace2021refrec} introduced a point cloud-specific self-training strategy with pseudo label refinement.

\subsection{Domain adaptive pose estimation}
\label{Sec:DA_pose_estimation}
Many of the introduced concepts for domain adaptation were adapted to general human/animal pose estimation and clinical human pose estimation.
\citet{martinez2018investigating} performed adversarial feature alignment for 2D human pose estimation from depth maps.
\citet{liu2022adapted} proposed semantically aware feature alignment coupled with a skeleton-aware pose refinement module for 3D human pose estimation from RGB images.
\citet{yang20183d} addressed the same task through adversarial output adaptation.
\citet{cao2019cross,li2021synthetic,mu2020learning} suggested different forms of self-training for 2D animal pose estimation.
\citet{kim2022unified} proposed a multi-level adaptation method for 2D human pose estimation, comprising style transfer at the input level and self-training with the Mean Teacher at the output level.
In the clinical context, \citet{srivastav2022unsupervised} presented a self-training framework with domain-specific normalization layers \citep{chang2019domain} for 2D clinician pose estimation and instance segmentation in the operating room.
Two multi-level adaptation strategies for 2D in-bed pose estimation, adapting from uncovered to covered patients on thermal images, were presented by \citet{afham2022towards,chi2022multi}.
The authors combined image-to-image translation at the input level with extreme augmentations and knowledge distillation \citep{afham2022towards} and with adversarial feature alignment and self-training \citep{chi2022multi}, respectively.

Compared to all discussed works, our method includes three essential methodical novelties.
First, it is the first approach to domain adaptive human pose estimation from 3D point clouds.
Second, unlike guiding self-training with pseudo labels through uncertainty estimates, we filter pseudo labels based on plausibility constraints derived from prior knowledge about the output space distribution.
Third, unlike adversarial and self-training-based output adaptation, we perform output space adaptation through anatomy-constrained optimization, realized by embedding anatomical constraints into a loss function.
The latter contribution is technically related to constrained optimization for medical image segmentation, introduced by \citet{kervadec2019constrained} for weakly-supervised learning and adapted to domain adaptation by \citet{bateson2021constrained}.
However, their proposed constraints on the sizes of target structures do not apply to human pose estimation, which requires specifically tailored constraints on the human skeleton graph.
Few works used such anatomical losses for 3D human pose estimation.
A geometric constraint on the ratio of bone lengths was proposed by \citet{zhang2020weakly,zhou2017towards} to regularize supervised learning with weak 2D pose ground truth.
Moreover, \citet{cao2020anatomy,sun2017compositional} introduced bone and symmetry losses as additional penalties in a fully supervised setting, where accurate ground truth poses, including precise bone lengths, are available.
These scenarios are substantially different from our unsupervised setting, where the anatomical loss functions are the only source of supervision on unlabeled target data and are derived from weaker constraints.

\section{Methods}
\subsection{Problem setup and notation}
Point cloud-based 3D human pose estimation aims at predicting the 3D positions of $K$ human joints of interest, $\boldsymbol{Y}\in\mathbb{R}^{K\times 3}$, from a 3D input point cloud $\boldsymbol{X}\in\mathbb{R}^{N\times 3}$.
We address the task in a domain adaptation setting, where training data consists of a labeled source dataset $\mathcal{S}=\{(\boldsymbol{X}_s,\boldsymbol{Y}_s)\}_{s=1}^{|\mathcal{S}|}$ and a shifted unlabeled target dataset $\mathcal{T}=\{\boldsymbol{X}_t\}_{t=1}^{|\mathcal{T}|}$.
The goal is to learn a function $f$ with parameters $\boldsymbol{\theta}_f$ that predicts human poses as $\boldsymbol{\hat{Y}}=f(\boldsymbol{X};\boldsymbol{\theta}_f)$ and achieves optimal performance on target data at test time.
We aim to solve the problem both in the UDA and SFDA setting.
UDA assumes simultaneous access to source and target data.
In SFDA, by contrast, source and target data are only accessible in successive stages.
The model is initially trained on source data and subsequently adapted to unlabeled target data without access to source data.

\textbf{Notation.} 
For a human pose $\boldsymbol{Y}$, we indicate individual joints as $\boldsymbol{y}_k\in\mathbb{R}^3$ and treat them as the nodes of a skeleton graph.
We denote $\mathcal{B}=\{\boldsymbol{b}_i\}_{i=1}^{N_{\beta}}$ as the set of all bone vectors $\boldsymbol{b}_i\in\mathbb{R}^3$ that connect two joints in the skeleton graph, and $\boldsymbol{b}_{t,i}$ indicates the i-th bone vector of the indexed pose $\boldsymbol{Y}_t$.
We further indicate $\mathcal{B}_{\lambda}\subset\mathcal{B}$ as the subset of $N_{\lambda}$ bones $\boldsymbol{b}_i^{\lambda}$ of the left body side that have a counterpart $\boldsymbol{b}_i^{\rho}\in\mathcal{B}_{\rho}$ on the right body side.
Finally, we term $\mathcal{B}_{\zeta}=\{(\boldsymbol{b}_i,\boldsymbol{b}_j)\}$ as the set of all $N_{\zeta}$ pairs of bone vectors that are connected by a joint and define $\mathcal{I}_{\zeta}=\{(i,j)\}$ as the corresponding set of indices.

\subsection{Overview}
\begin{figure*}[t]
\begin{center}
  \includegraphics[width=\linewidth]{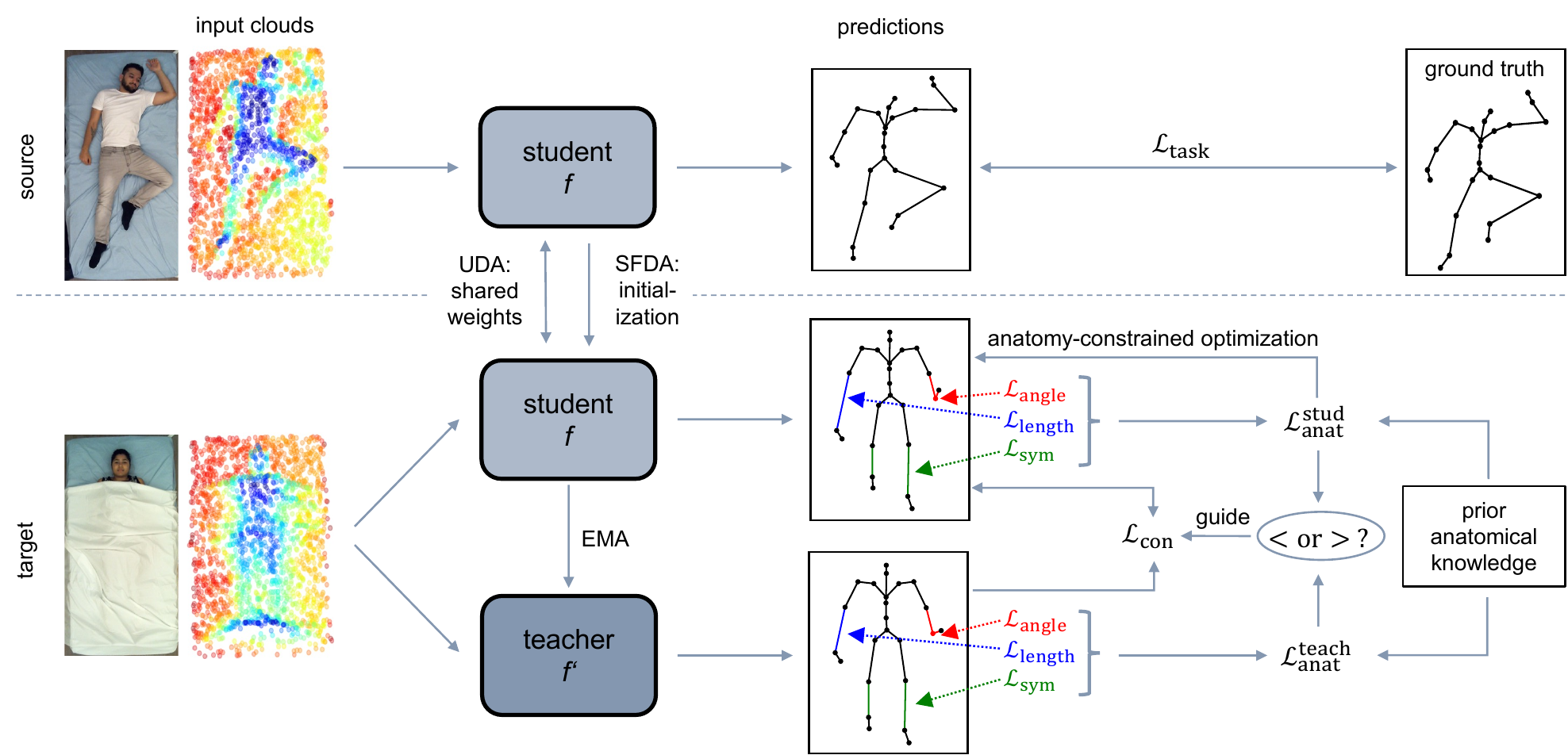}
\end{center}
  \caption{
  Overview of our method for domain adaptive human pose estimation from point clouds (RGB images are only shown for better visualization).
  The framework comprises a learning student model and a teacher model, which represents the exponential moving average (EMA) of the student.
  While source training of the student consists in minimizing a supervised task loss, we perform anatomy-guided learning in the unlabeled target domain.
  Based on prior knowledge about human anatomy, we formulate an anatomical loss that measures the violation of \textcolor{mygreen}{symmetry}, \textcolor{myblue}{bone lengths}, and \textcolor{myred}{joint angle} constraints.
  We use the loss to 1) explicitly constrain the student predictions to the space of plausible human poses and 2) filter pseudo labels from the teacher network for self-training according to their anatomical plausibility.
  As such, the method is applicable to unsupervised domain adaptation (UDA), where the model is jointly trained on the source and target data, and source-free domain adaptation (SFDA), which accesses the domains in two successive steps.
  }
\label{Fig:method_overview}
\end{figure*}

An overview of our proposed method to solve the above problem is shown in Fig.~\ref{Fig:method_overview}.
While supervised learning on labeled source data is performed by minimizing the task loss
\begin{equation}
    \mathcal{L}_{\mathrm{task}}(\boldsymbol{\theta}_f;\mathcal{S})=\frac{1}{|\mathcal{S}|}\sum_{s}\frac{1}{K}\left\|\boldsymbol{Y}_{s}-\hat{\boldsymbol{Y}}_{s}\right\|_1
\end{equation}
we aim to bridge the domain gap by exploiting domain-invariant prior knowledge about human anatomy.
To this end, we introduce two complementary anatomy-based training strategies that guide the learning process in the unlabeled target domain.
On the one hand, we directly embed the prior knowledge into an anatomical loss function ($\mathcal{L}_{\mathrm{anat}}$) to penalize anatomically implausible predictions.
We derive the loss from an anatomically constrained optimization problem in Sec.~\ref{Sec:anat-cons-opt}.
On the other hand, we leverage prior anatomical knowledge to filter pseudo labels for self-training with the Mean Teacher, realized by $\mathcal{L}_{\mathrm{con}}$ (see Sec.~\ref{Sec:mean-teacher} for details).

\subsection{Anatomy-constrained optimization}
\label{Sec:anat-cons-opt}
We start our discussion for UDA.
Our goal is to guide the learning on unlabeled target data by constraining predictions to the space of anatomically plausible poses.
To this end, we formulate network training as the constrained optimization problem
\begin{equation}
\begin{split}
    \min_{\boldsymbol{\theta}_f}\quad&\mathcal{L}_{\mathrm{task}}(\boldsymbol{\theta}_f;\mathcal{S})\\
    \mathrm{s.t.}\quad &\hat{\boldsymbol{Y}}_t\; \text{is a plausible human pose}\quad t=1,...,|\mathcal{T}|
\end{split}
\end{equation}
At this stage, the essential question is how to formalize the plausibility constraint.
Given the high complexity of the human pose space, we approximate it by means of explicit prior knowledge about human anatomy. Specifically, we combine three simpler constraints on the human skeleton graph that are strong indicators for the plausibility of a pose:
\begin{itemize}
    
    \item \textbf{Symmetric limbs:}
    Corresponding limb pairs ($\boldsymbol{b}_i^{\lambda},\boldsymbol{b}_i^{\rho}$) of the human body typically have roughly equal lengths, with a deviation $\bigl|\|\boldsymbol{b}_i^{\lambda}\|_2-\|\boldsymbol{b}_i^{\rho}\|_2\bigr|<\delta_i$ smaller than a limb-specific tolerance $\delta_i$.
    We set $\delta_i=0$ by default but retain the option for an adjustment when dealing with pathologically asymmetric limbs.
    
    \item \textbf{Plausible bone lengths:}
    The lengths of human bones $\boldsymbol{b}_i$ are constrained by bone-specific upper and lower bounds $u_i^{\beta}$ and $l_i^{\beta}$, i.e., $l_i^{\beta}\leq\|\boldsymbol{b}_i\|_2\leq u_i^{\beta}$.
    Precise values for $u_i^{\beta}$ and $l_i^{\beta}$ can be looked up in an anatomical textbook or inferred from the statistics of the training set.
    
    \item \textbf{Plausible joint angles:} 
    Human joints cannot freely rotate in 3D space but the range of angles that can be taken is limited.
    More formally, the normalized dot product of two connected bone vectors $(\boldsymbol{b}_i,\boldsymbol{b}_j)\in\mathcal{B}_{\zeta}$ is constrained by joint-specific upper and lower bounds $u_{ij}^{\alpha}$ and $l_{ij}^{\alpha}$, i.e., $l_{ij}^{\alpha}\leq\boldsymbol{b}_i/\|\boldsymbol{b}_i\|_2\cdot\boldsymbol{b}_j/\|\boldsymbol{b}_j\|_2\leq u_{ij}^{\alpha}$.
    Again, the precise determination of upper and lower bounds can be based on an anatomical textbook or the statistics of the training set.

\end{itemize}
Altogether, this yields the novel optimization problem
\begin{equation}
\label{Eq:constrained_optimization_problem}
\begin{split}
    \min_{\boldsymbol{\theta}_f}\;\mathcal{L}&_{\mathrm{task}}(\boldsymbol{\theta}_f;\mathcal{S})\\
    \mathrm{s.t.}\;-&\delta_i<\|\boldsymbol{b}_{t,i}^{\lambda}\|_2-\|\boldsymbol{b}_{t,i}^{\rho}\|_2<\delta_i \quad \; \; \, i=1,...,N_{\lambda};t=1,...,|\mathcal{T}|\\
    &l_i^{\beta}\leq\|\boldsymbol{b}_{t,i}\|_2\leq u_i^{\beta} \qquad \qquad \quad \,i=1,...,N_{\beta};t=1,...,|\mathcal{T}|\\
    &l_{ij}^{\alpha}\leq\frac{\boldsymbol{b}_{t,i}}{\|\boldsymbol{b}_{t,i}\|_2}\cdot\frac{\boldsymbol{b}_{t,j}}{\|\boldsymbol{b}_{t,j}\|_2}\leq u_{ij}^{\alpha}\quad \;\forall (i,j)\in\mathcal{I}_{\zeta};t=1,...,|\mathcal{T}|\\
\end{split}
\end{equation}
As discussed in prior work \citep{bateson2021constrained,kervadec2019constrained}, a known method to solve such a problem requires the minimization of the Lagrangian dual \citep{bertsekas1997nonlinear}.
However, this technique becomes unstable and computationally intractable when deep neural networks are involved.
Alternatively, the problem can be approximated by relaxing the hard constraints to soft constraints in the form of differentiable loss functions that augment the original objective and penalize violations of the constraints.
To implement this, we define the base penalty function
\begin{equation}
\label{eq:base_penalty}
    \ell(x;l,u)=\begin{cases}
        |x-l| & x < l \\
        |x-u| & x > u \\
        0 & l < x < u
        \end{cases}
\end{equation}
which outputs 0 if the input $x$ lies inside the lower and upper bounds and penalizes inputs outside this range with a linear L1 loss.
We also experimented with a quadratic penalty, which performed slightly worse (Sec.~\ref{Sec:Ablation_anatomical}).
Given a human pose $\boldsymbol{Y}$, the violation of our anatomical constraints is then penalized by the loss functions
\begin{equation}
\label{Eq:anatmical_losses_individual}
\begin{split}
    &\mathcal{L}_{\mathrm{sym}}(\boldsymbol{Y})=\frac{1}{N_{\lambda}}\sum_{i=1}^{N_{\lambda}}\ell\left(\|\boldsymbol{b}_{i}^{\lambda}\|_2-\|\boldsymbol{b}_{i}^{\rho}\|_2,-\delta_i, \delta_i\right)\\
    &\mathcal{L}_{\mathrm{\textcolor{rev_changes}{length}}}(\boldsymbol{Y})=\frac{1}{N_{\beta}}\sum_{i=1}^{N_{\beta}}\ell\left(\|\boldsymbol{b}_{i}\|_2,l_i^{\beta}, u_i^{\beta}\right)\\
    &\mathcal{L}_{\mathrm{angle}}(\boldsymbol{Y})=\frac{1}{N_{\zeta}}\sum_{(i,j)\in\mathcal{I}_{\zeta}}\ell\left(\frac{\boldsymbol{b}_{i}}{\|\boldsymbol{b}_{i}\|_2}\cdot\frac{\boldsymbol{b}_{j}}{\|\boldsymbol{b}_{j}\|_2};l_{ij}^{\alpha},u_{ij}^{\alpha}\right)
\end{split}
\end{equation}
This enables us to replace the constrained optimization problem \textcolor{rev_changes}{in Eq.~\eqref{Eq:constrained_optimization_problem}} by the standard minimization of the joint loss function
\begin{equation}
\label{Eq:loss-task-anat}
    \mathcal{L}(\boldsymbol{\theta}_f;\mathcal{S},\mathcal{T})=\mathcal{L}_{\mathrm{task}}(\boldsymbol{\theta}_f;\mathcal{S})+\lambda_1\mathcal{L}_{\mathrm{anat}}(\boldsymbol{\theta}_f;\mathcal{T})
\end{equation}
with the anatomical loss
\begin{equation}
    \mathcal{L}_{\mathrm{anat}}(\boldsymbol{\theta}_f,\mathcal{T})=\frac{1}{|\mathcal{T}|}\sum_{t}\Bigl[\mathcal{L}_{\mathrm{sym}}(\hat{\boldsymbol{Y}}_t)+\mathcal{L}_{\mathrm{\textcolor{rev_changes}{length}}}(\hat{\boldsymbol{Y}}_t)+\mathcal{L}_{\mathrm{angle}}(\hat{\boldsymbol{Y}}_t)\Bigr]
\end{equation}
and the weighting factor $\lambda_1$.
Individual weighting factors for each loss were explored but did not yield an improvement.

\subsubsection{Optimization and SFDA}
\label{Sec:anat-optimization}
Since $\mathcal{L}_{\mathrm{task}}(\mathcal{S})$ and $\mathcal{L}_{\mathrm{anat}}(\mathcal{T})$ each only depend on a single domain, the above method is technically also applicable to SFDA by separately minimizing the two losses in successive stages.
(Strictly speaking, when deriving upper and lower bounds from the training set, the anatomical loss still accesses labels from the source domain.
But unlike visual input data, the upper and lower bounds of the label distribution do not represent sensitive information in terms of data protection, and sharing them among institutions is uncritical.)
However, for both UDA and SFDA, when minimizing $\mathcal{L}_{\mathrm{anat}}(\mathcal{T})$ over all model parameters $\boldsymbol{\theta}_f$, we observed a mode collapse in the target domain, where the model predicted a roughly fixed anatomically plausible pose independent of the input.
The phenomenon was particularly prominent in SFDA as the absence of joint supervision on source data caused the model to forget that the predicted pose should match the given input.
As suggested in our preceding work \citep{bigalke2021domain}, an intuitive solution to this problem is to minimize $\mathcal{L}_{\mathrm{anat}}(\mathcal{T})$ over a restricted subset of network parameters $\boldsymbol{\theta}_g\subset\boldsymbol{\theta}_f$ while minimizing $\mathcal{L}_{\mathrm{task}}$ over all parameters.
We experimentally found that only optimizing the feature extractor $g$ of $f$ yields excellent results in UDA, whereas SFDA required a further restriction to the parameters of the BatchNorm layers of $g$ to achieve decent results.
While this technique successfully prevents the mode collapse, it also limits the adaptation capacity of the network.
As an alternative, we therefore propose to combine anatomy-constrained optimization with supervision through pseudo labels, which can prevent the mode collapse without restricting the adaptability of the network.
In our prior work, we already experimentally demonstrated that anatomy-constrained optimization works particularly well in combination with pseudo labels provided by the Mean Teacher \citep{french2017self}.
In the following Sec.~\ref{Sec:mean-teacher}, we formalize the Mean Teacher framework in the context of our problem and extend the standard version by filtering the provided pseudo labels according to their anatomical plausibility.

\subsection{Self-training with the Mean Teacher}
\label{Sec:mean-teacher}
The Mean Teacher framework \citep{french2017self,tarvainen2017mean} extends the learning model $f$, from now on denoted as the student model, by a second so-called teacher model $f'$ with identical architecture.
Unlike the student model, the weights of the teacher $\boldsymbol{\theta}'_f$ are not optimized by gradient descent but given as the exponential moving average (EMA) of the student's weights, updated as
\begin{equation}
    \boldsymbol{\theta}'_{f,i}=\mu\boldsymbol{\theta}'_{f,i-1}+(1-\mu)\boldsymbol{\theta}_{f,i}
\end{equation}
at iteration $i$ with momentum $\mu$.
Thus, the teacher can be seen as a temporal ensemble of the student and is therefore expected to provide---on average---more stable and accurate predictions than the student.
The essential idea of the framework is to leverage this superiority of the teacher by supervising student predictions on unlabeled target data with pseudo labels provided by the teacher.
This is implemented by a consistency loss
\begin{equation}
\label{Eq:cons-loss-base}
    \mathcal{L}_{\mathrm{con}}(\boldsymbol{\theta}_f;\boldsymbol{\theta}'_f,\mathcal{T})=\frac{1}{|\mathcal{T}|}\sum_{t}\frac{1}{K}\left\|\hat{\boldsymbol{Y}}_{t}-\hat{\boldsymbol{Y}}'_{t}\right\|_1
\end{equation}
encouraging predictions $\hat{\boldsymbol{Y}}'_{t}=f'(\boldsymbol{X}_t;\boldsymbol{\theta}'_{f})$ by the teacher and $\hat{\boldsymbol{Y}}_{t}=f(\boldsymbol{X}_t;\boldsymbol{\theta}_{f})$ by the student to be consistent.
To prevent trivial solutions and vanishing gradients, teacher and student operate on different augmentations of the same input sample that are reversed in the output space to align the predicted poses.
In our point cloud-based framework, augmentations consist of random global translation, rotation, and subsampling of input points.

\subsubsection{Anatomy-guided filtering of pseudo labels}
For the consistency loss to efficiently guide the learning process on target data, predictions by the teacher should be more accurate than those of the student.
While this is expected on average, there will be samples where the teacher prediction is inferior to the student prediction.
In such cases, the consistency loss in \textcolor{rev_changes}{Eq.~\eqref{Eq:cons-loss-base}} drives the student towards a worse solution and thus hampers the learning process.
Instead, we would ideally filter the pseudo labels provided by the teacher and only use those labels for supervision that are more accurate than the current predictions of the student.
Since accuracy itself can obviously not be measured in the absence of ground truth, another criterion for filtering pseudo labels is needed.

We propose to filter pseudo labels based on their anatomical plausibility.
Specifically, we argue that anatomically plausible poses are more likely to be correct than implausible poses.
Consequently, we assess pseudo labels by the teacher and predictions by the student by measuring their plausibility with our three anatomical loss functions in \textcolor{rev_changes}{Eq.~\eqref{Eq:anatmical_losses_individual}}.
Given the comparisons of the three loss functions, we use only those pseudo labels for supervision, for which at least two out of three anatomical losses indicate a higher plausibility (smaller value) than for the corresponding student predictions.
Note that we could alternatively select pseudo labels by comparing the sum of all three losses ($\mathcal{L}_{\mathrm{anat}}$) or just a single loss, but the above criterion gave the best results in the ablation study (Sec.~\ref{Sec:Ablation_filtering}).

To formalize the approach, we define the boolean function \textcolor{rev_changes}{$\mathbbm{1}(condition)$}, which is equal to 1 if the condition is fulfilled and 0 otherwise.
Given teacher and student predictions $\hat{\boldsymbol{Y}}'$ and $\hat{\boldsymbol{Y}}$, we then define the function
\textcolor{rev_changes}{
\begin{equation}
\label{Eq:filter-criterion}
\begin{split}
    h(\hat{\boldsymbol{Y}}',\hat{\boldsymbol{Y}})=\mathbbm{1}\biggl(\Bigl[&\mathbbm{1}\left(\mathcal{L}_{\mathrm{sym}}(\hat{\boldsymbol{Y}}')<\mathcal{L}_{\mathrm{sym}}(\hat{\boldsymbol{Y}})\right)\\
    +&\mathbbm{1}\left(\mathcal{L}_{\mathrm{length}}(\hat{\boldsymbol{Y}}')<\mathcal{L}_{\mathrm{length}}(\hat{\boldsymbol{Y}})\right)\\
    +&\mathbbm{1}\left(\mathcal{L}_{\mathrm{angle}}(\hat{\boldsymbol{Y}}')<\mathcal{L}_{\mathrm{angle}}(\hat{\boldsymbol{Y}})\right)\Bigr]\geq 2\biggr)
\end{split}
\end{equation}
}which outputs 1 if our criterion affirms the use of the teacher prediction for supervision and 0 otherwise.
We finally reformulate the consistency loss from \textcolor{rev_changes}{Eq.~\eqref{Eq:cons-loss-base}} as
\begin{equation}
    \mathcal{L}_{\mathrm{con}}(\boldsymbol{\theta}_f;\boldsymbol{\theta}'_f,\mathcal{T})=\frac{1}{|\mathcal{T}|}\sum_{t}\frac{1}{K}h(\hat{\boldsymbol{Y}}',\hat{\boldsymbol{Y}})\cdot\left\|\hat{\boldsymbol{Y}}_{t}-\hat{\boldsymbol{Y}}'_{t}\right\|_1
\end{equation}

Taking altogether, we integrate this consistency loss into our previous objective function \textcolor{rev_changes}{from Eq.~\eqref{Eq:loss-task-anat}}. To perform UDA, we thus minimize
\begin{equation}
\label{eq:overall_loss}
\textcolor{rev_changes}{
\begin{split}
    \mathcal{L}(\boldsymbol{\theta}_f;\boldsymbol{\theta}'_f,\mathcal{S},\mathcal{T})=&\mathcal{L}_{\mathrm{task}}(\boldsymbol{\theta}_f;\mathcal{S})\\
    &+\lambda(\tau)\lambda_1\mathcal{L}_{\mathrm{anat}}(\boldsymbol{\theta}_f;\mathcal{T})\\
    &+\lambda(\tau)\lambda_2\mathcal{L}_{\mathrm{con}}(\boldsymbol{\theta}_f;\boldsymbol{\theta}'_f,\mathcal{T})
\end{split}}
\end{equation}
Here, $\lambda(\tau)=\mathrm{exp}(-5(1-\mathrm{min}(\tau/T, 1)^2))$ depends on the current epoch $\tau$ and continually increases from 0 to 1 during the first $T$ epochs, as suggested by \citet{tarvainen2017mean}, while $\lambda_2$ is a fixed weighting factor.
Time dependency is needed to suppress noisy gradients from $\mathcal{L}_{\mathrm{con}}$ and $\mathcal{L}_{\mathrm{anat}}$ at early epochs when the weights of the student and the teacher model are still close to initialization.

For SFDA, we adapt the model pre-trained on source data by minimizing
\begin{equation}
\label{Eq:overall-loss-sfda}
    \mathcal{L}(\boldsymbol{\theta}_f;\boldsymbol{\theta}'_f,\mathcal{T})=\lambda_1\mathcal{L}_{\mathrm{anat}}(\boldsymbol{\theta}_f;\mathcal{T})+\lambda_2\mathcal{L}_{\mathrm{con}}(\boldsymbol{\theta}_f;\boldsymbol{\theta}'_f,\mathcal{T})
\end{equation}
Time-dependent weighting is not required because pre-training avoids noisy gradients.
Note that, for SFDA, the student and the teacher are initialized with the same weights of the pre-trained source model.
This is in contrast to the initialization with different random weights in UDA.
Furthermore, related to our discussion in Sec.~\ref{Sec:anat-optimization}, we found it beneficial to minimize the loss in \textcolor{rev_changes}{Eq.~\eqref{Eq:overall-loss-sfda}} just with respect to the weights of the feature extractor of $f$ while freezing the network heads.
This reduces the risk of the model forgetting source knowledge, which is constantly present when dealing with SFDA.

\subsection{Point cloud-based 3D pose estimation}
While our formulation is agnostic to the specific implementation of the function $f$, we realize point cloud-based 3D pose estimation as follows.
Given an input point cloud $\boldsymbol{X}\in\mathbb{R}^{N\times 3}$, we estimate the associated 3D pose $\boldsymbol{\hat{Y}}\in\mathbb{R}^{K\times 3}$ as the weighted sum over the $N$ input points $\boldsymbol{x}_{i}\in\mathbb{R}^3$.
To this end, we design $f$ to output a stack of $K$ softmax-normalized weight maps $\boldsymbol{W}=f(\boldsymbol{X};\boldsymbol{\theta_f})\in\mathbb{R}^{N\times K}$ over the input points.
The $k$-th predicted joint is then given by $\boldsymbol{\hat{y}}_k=\sum_{i=1}^{N}\boldsymbol{x}_i\cdot w_{ik}$.
In our work, we implement $f$ as the segmentation architecture of DGCNN \citep{wang2019dynamic} with 40 neighbors in the neighborhood graph.
The network comprises a feature extractor with six convolutional layers and network heads with a shared MLP of three fully-connected layers, yielding 986k model parameters.

\section{Experimental setup}
\subsection{Datasets}
We evaluate our method for the use case of in-bed patient monitoring, using two in-bed human pose datasets: the public SLP dataset \citep{liu2019seeing,liu2022simultaneously} and an in-house dataset denoted as MVIBP (multi-view in-bed pose) dataset.

\paragraph{SLP}
The SLP dataset comprises single-view depth frames of 109 subjects, captured with a Kinect v2 mounted centrally above the bed.
Each subject takes 45 arbitrary resting poses, evenly distributed across supine and lateral (left, right) positions.
For each pose, the subjects do not move until three frames with varying cover conditions (no cover, thin cover $\sim$\SI{1}{mm}, thick cover $\sim$\SI{3}{mm}) are captured.
That way, pose annotations for frames without a cover are also valid for frames with cover.
While the original dataset includes 2D joints, \citet{clever2022bodypressure} provided the 24 joints of the SMPL model \citep{loper2015smpl} as 3D ground truth for the first 102 subjects.
We restrict our experiments to these subjects.
The first 70 subjects are used for training, subjects 71-80 for validation, and subjects 81-102 for testing.
As pre-processing, we transformed depth frames to point clouds using the internal camera parameters and removed all points outside a predefined box around the bed.

\paragraph{MVIBP}
The MVIBP dataset comprises multi-view depth frames of 13 subjects captured by three synchronized Azure Kinect cameras on the left and right sides and at the foot of the bed.
We recorded video data of the subjects, which were asked to freely move while staying in either supine, left, or right position\footnote{The conduct of our study was approved by the ethical review board of L\"ubeck University.
Only healthy adults were included, and all subjects gave their informed consent.}.
Subjects remained permanently uncovered, but---contrary to the SLP dataset---we occasionally bedded them on a small or large pillow.
To further simulate a clinically realistic scenario, we used positioning aids, and subjects sometimes wore a respiratory mask (\textit{not} used for active ventilation).
Given the video data, we extracted discrete frames at fixed time intervals.
After removing visually similar frames, we processed the remaining ones in four steps.
First, we transformed the depth frames from all three cameras to a point cloud using the internal camera parameters.
Second, using the external calibration among the cameras, we rotated each cloud to world coordinates and merged the three clouds.
Third, we removed all points outside a predefined box around the bed.
Fourth, we downsampled the cloud with a voxel filter with an edge length of \SI{2}{cm}.
For each resulting cloud, we manually annotated the ground truth positions of ten joints (feet, knees, shoulders, elbows, and hands) according to the location of the corresponding SMPL joints.
To eliminate duplicate poses from the dataset, we only kept those frames where at least one joint moved by more than a threshold of \SI{10}{cm} compared to the previous extracted frame.
This results in a total of 2408 frames, 1165 showing a supine and 1243 a lateral position.
Regarding the data split, we use three subjects (361 frames, 177 with supine and 184 with lateral position) for testing and the remaining subjects for training.
A validation set is not required because hyper-parameters are not tuned on this dataset.

\subsection{Adaptation scenarios}
Given the two datasets, we consider two adaptation scenarios, featuring domain shifts with different characteristics.

\paragraph{Uncover$\rightarrow$cover}
Using only the SLP dataset, we consider uncovered subjects as the labeled source and covered subjects as the unlabeled target domain.
Thus, the domain shift consists in the occlusion of the subjects by a cover.
The scenario is relevant in practical applications because the annotation of uncovered subjects is viable, while it is virtually infeasible for covered patients in practice.
(The same adaptation problem for thermal image data was addressed in the IEEE VIP Cup 2021 \citep{liu2022privacy}.)
For our experiments, we randomly divide the training data by subject into three splits with 30, 20, and 20 subjects.
For each split, we use only one cover condition---uncover, thin cover, and thick cover, respectively---while the remaining data is discarded.
This yields 30 subjects as the source and 40 subjects as the target domain.
For validation and test set, we use both the thin and the thick cover for all frames of all subjects.

\paragraph{SLP$\rightarrow$MVIBP}
We focus on uncovered subjects and consider SLP as the labeled source and MVIBP as the unlabeled target dataset.
The domain shift results from a broad range of factors:
1) different sensors (Kinect v2 vs.~Azure Kinect), 2) different camera perspectives and camera-to-bed distances (yielding differing distributions of points in 3D space), 3) different geometry of the used beds (the bed in MVIBP has a headboard), 4) pillows, positioning aids, and respiration masks are only used in MVIBP, 5) cropped point clouds from MVIBP may contain persons walking around the bed.
This scenario is relevant in clinical practice as it simulates the deployment of a model in a different environment, e.g., in another hospital.
In our experiments, we use the training set from the SLP dataset (70 uncovered subjects) as the labeled source dataset and the training set from MVIBP (10 subjects) as the unlabeled target dataset.
Results are reported on the test set of MVIBP (3 subjects).
Since the annotated pose skeletons in the two dataset are not identical (see Fig.~\ref{Fig:motivation}, right), we restrict the evaluation to the matching joint pairs, namely feet, knees, shoulders, elbows, and hands.

\subsection{Implementation details}
We implement our method in PyTorch and use the Adam optimizer for training.
We train for 100 epochs for UDA and for 80 epochs for SFDA with a constant learning rate of 0.001.
Batches are composed of 8 source and 8 target samples for UDA and of 8 target samples only for SFDA.
The weighting factors in \textcolor{rev_changes}{Eq.~\eqref{eq:overall_loss}} are set to $\lambda_1=0.1$ and $\lambda_2=1$, and the ramp-up length $T$ is set to 40 epochs.
The momentum $\mu$ for updating the teacher's weights is set to 0.99 for UDA and 0.9996 for SFDA.
Upper and lower bounds $u_{ij}^{\alpha}$, $u_i^{\beta}$ / $l_{ij}^{\alpha}$, $l_i^{\beta}$ of our anatomical constraints are set to the max/min values from the training set of the source domain.
For regularization, we use a weight decay of 1e-5 and augment the input point clouds by random rotation around the z-axis, translation, and subsampling to 2048 points.
For further details, we refer to our public code at \url{https://github.com/multimodallearning/da-3dhpe-anatomy}.
The above hyper-parameters of our method and the hyper-parameters of all comparison methods (Sec.~\ref{Sec:comparison_methods}) were tuned on the validation set of the target domain under the uncover$\rightarrow$cover scenario and kept fix for adaptation from SLP to MVIBP.
Final results are reported on the test sets of the target domain in terms of the mean per joint position error (MPJPE).

\subsection{Comparison methods}
\label{Sec:comparison_methods}
In this section, we describe the comparison methods used in the experiments.
We start by describing the lower and upper bounds.

\textit{1) Mean pose.}
For each sample from the test set, we estimate the pose as the mean pose over all training samples.
To construct this mean pose, we anchor the root joint of all training poses at the origin and compute the mean over these centered poses.
For evaluation, we apply the same anchoring to the test pose and then compare it to the mean pose.
We use this trivial baseline to assess the variability of the used datasets.
Note, however, that this baseline accesses ground truth information (location of root joint) at inference time.

\textit{2) Source-only.}
The source-only model is exclusively trained on labeled source data without adaptation techniques and represents a lower bound.

\textit{3) Target-only.}
The target-only model (oracle) is trained on labeled data from the target domain and thus constitutes an upper bound.

To our knowledge, there is no prior work for domain adaptive 3D human pose estimation from point clouds.
Therefore, we adapt a comprehensive set of state-of-the-art DA methods to the problem.
We primarily describe UDA methods.

\textit{4) MMD.} 
Similar to the methods by \citet{rozantsev2018beyond,tzeng2014deep}, the distributions of source and target features are aligned by minimizing the Maximum Mean Discrepancy (MMD) loss \citep{gretton2006kernel}, computed for the global feature vector after conv6 in the DGCNN.
We explored a linear and an exponential kernel, with the former yielding slightly better results.

\textit{5) DANN.}
\citet{ganin2015unsupervised} proposed to learn domain-invariant features by adversarial learning: a domain discriminator learns to distinguish source and target features while the feature extractor is trained to fool the discriminator.
Adversarial optimization is realized by a gradient reversal layer after the feature extractor.
We implement the discriminator as a fully-connected network with three layers and apply it to the global feature vector after conv6 in the DGCNN.

\textit{6) DefRec.}
The method by \citet{achituve2021self} performs point cloud-based DA through self-supervised learning.
The pretext task is to reconstruct the original input point cloud from a deformed version, where a subset of points is replaced by new points sampled from an isotropic Gaussian distribution with small standard deviation.

\textit{7) SSDispPred.}
Inspired by the method of \citet{doersch2015unsupervised}, we design a novel pretext task for self-supervised DA, which consists in predicting the displacement vector between two randomly sampled patches from an input cloud.

\textit{8) AdvOutAdapt.}
We adopt the adversarial output adaptation method by \citet{yang20183d}.
A discriminator learns to distinguish predicted poses on target data from ground truth poses in the source domain.
Meanwhile, the pose estimation network is trained to fool the discriminator by predicting poses that match the distribution of ground truth poses.
As for the implementation of the discriminator, we explored diverse architectures of fully-connected and graph neural networks, with the former yielding better results.
This method is related to our anatomy-constrained optimization since the discriminator could theoretically learn to penalize implausible predictions similar to our anatomical losses.

\textit{9) CC-SSL.} \citet{mu2020learning} proposed a consistency-constrained curriculum learning strategy for efficient self-training with pseudo labels.
First, the confidence for initial pseudo labels from the source-only model is assessed by measuring the consistency under input perturbations.
The most confident pseudo labels are then selected for supervised training.
After some epochs, the pseudo labels are updated, their confidence is reassessed, and a larger proportion of pseudo labels is selected for the next stage of supervised training.
This procedure is repeated several times.

\textit{10) MCD.}
Inspired by the concept of Maximum Classifier Discrepancy \citep{saito2018maximum}, we extend the pose estimation model by a second network head with a different weight initialization.
DA is realized by performing two sequential optimization steps at each iteration.
First, the feature extractor and the network heads are jointly optimized on labeled source data.
Second, the feature extractor only is optimized on unlabeled target data by minimizing the discrepancy between the predictions of the network heads for the same input sample.

\textit{11) Mean Teacher.}
An extension of the Mean Teacher \citep{french2017self,srivastav2022unsupervised} is already part of our method (Sec.~\ref{Sec:mean-teacher}).
The original Mean Teacher thus corresponds to an ablated version of our method, excluding anatomy-guided filtering of pseudo labels and anatomy-constrained optimization.

We further describe three state-of-the-art comparison methods for SFDA.

\textit{12) UBNA.}
\citet{klingner2022unsupervised} perform SFDA by partially adapting the statistics of the BatchNorm layers to the target domain.
The authors use an exponentially decaying momentum factor for the adaptation such that the updated statistics represent a mix of the statistics from the source and target domain.

\textit{12) BNAdapt.}
\citet{zhang2020generalizable} also tackled SFDA by adapting the statistics of the BatchNorm layers.
Specifically, they proposed to use the statistics of the test batch itself at inference time instead of the running mean and variance captured during training.
Note, however, that this requires sufficiently large batches at test time, which are not always available.
We found a batch size of 64 to be a good trade-off between memory consumption and performance.
To minimize random effects due to the composition of the test batches, we repeat each experiment five times and report average scores.

\textit{13) Mean Teacher.} 
\citet{wang2022continual} extended the Mean Teacher to continual test time adaptation, which is closely related to SFDA.
The authors proposed to improve the quality of the pseudo labels from the Mean Teacher by averaging over multiple predictions under different input augmentations. 
Moreover, they addressed catastrophic forgetting by stochastically resetting a small ratio of weights to the original pre-trained weights after each iteration.
In our experiments, however, neither augmentation-averaged pseudo labels nor stochastic weight restoration brought any benefits.
Therefore, we use the standard Mean Teacher with frozen network heads, which is identical to the ablated version of our method.

\section{Results}

\subsection{Ablation study}
We start by analyzing the two essential components of our method, namely anatomy-constrained optimization in Sec.~\ref{Sec:Ablation_anatomical} and anatomy-guided filtering of pseudo-labels in Sec.~\ref{Sec:Ablation_filtering}.
The ablation experiments are performed under the uncover$\rightarrow$cover setting on the SLP dataset.

\subsubsection{Anatomy-constrained optimization}
\label{Sec:Ablation_anatomical}
In the first ablation experiment, we examine the effectiveness of the proposed anatomical loss functions \textcolor{rev_changes}{from Eq.~\eqref{Eq:anatmical_losses_individual}}.
\begin{table}[t]
  \caption{\label{Tab:ablation_anatomical}Mean per joint position error (MPJPE) for domain adaption with different anatomical loss functions compared to the source-only and target-only models. For each of the three anatomical constraints, we compare a linear L1 against a quadratic L2 penalty. The evaluation is performed for UDA under the uncover$\rightarrow$cover adaptation scenario on the SLP dataset.}
  \centering
  \sisetup{detect-weight=true,detect-inline-weight=math}
  \begin{tabular}{lccS[table-format=3.1]}
  \toprule
  \bfseries Method & \bfseries L1 & \bfseries L2 & \bfseries {MPJPE [mm]}\\
  \midrule
  source-only & & & 130.4\\
  target-only & & & 67.7\\
  \midrule
  $\mathcal{L}_{\mathrm{angle}}$& \checkmark & & $\boldsymbol{106.7}$\\
  $\mathcal{L}_{\mathrm{angle}}$& & \checkmark & 119.3\\
  \midrule
  $\mathcal{L}_{\mathrm{sym}}$& \checkmark & & $\boldsymbol{105.9}$\\
  $\mathcal{L}_{\mathrm{sym}}$& & \checkmark & 108.6\\
  \midrule
  $\mathcal{L}_{\mathrm{\textcolor{rev_changes}{length}}}$& \checkmark & & $\boldsymbol{102.9}$\\
  $\mathcal{L}_{\mathrm{\textcolor{rev_changes}{length}}}$& & \checkmark & 104.1\\
  \midrule
  $\mathcal{L}_{\mathrm{anat}}$ & \checkmark & & \bfseries 96.6\\
  \bottomrule
  \end{tabular}
\end{table}
We consider the UDA setting, discard the Mean Teacher, and minimize $\mathcal{L}=\mathcal{L}_{\mathrm{task}}+\lambda_1\mathcal{L}_{\mathrm{x}}$ with $\mathcal{L}_{\mathrm{x}}\in\{\mathcal{L}_{\mathrm{sym}},\mathcal{L}_{\mathrm{angle}},\mathcal{L}_{\mathrm{\textcolor{rev_changes}{length}}},\mathcal{L}_{\mathrm{anat}}\}$.
For the three individual losses ($\mathcal{L}_{\mathrm{sym}},\mathcal{L}_{\mathrm{angle}},\mathcal{L}_{\mathrm{\textcolor{rev_changes}{length}}}$), we examine L1 and L2 penalties.

Results of the experiment are shown in Tab.~\ref{Tab:ablation_anatomical}.
Our insights are three-fold.
First, each of the three individual loss functions alone substantially reduces the error of the source-only baseline---irrespective of the used penalty function.
Second, for all three constraints, the L1 penalty is superior to the L2 penalty, whereby the gap is particularly notable for the angle constraint.
Third, aggregating the individual losses in $\mathcal{L}_{\mathrm{anat}}$ further improves performance.
This indicates that the three proposed constraints effectively complement each other, thus better approximating the space of plausible poses than any of the constraints alone.
Overall, our anatomy-constrained optimization reduces the error of the source-only model by 26\% and the gap between the source-only and the target-only model by 54\%.

\subsubsection{Anatomy guided filtering of pseudo labels}
\label{Sec:Ablation_filtering}
Next, we examine the effect of anatomy-guided filtering of pseudo labels.
We start by verifying our hypothesis that anatomically plausible pose estimates are more likely to be correct than implausible ones.
\begin{table}[t]
\setlength{\tabcolsep}{4pt}
  \caption{\label{Tab:ablation_mean-teacher}Mean per joint position error (MPJPE) for different techniques to filter pseudo labels under the Mean Teacher paradigm. The evaluation was performed for UDA and SFDA under the uncover$\rightarrow$cover adaptation scenario on the SLP dataset. Pearson coefficient R and corresponding significance value p indicate the correlation between the anatomical loss functions and the MPJPE, measured on predictions of the source-only model on the validation set of the target domain.}
  \centering
  \sisetup{detect-weight=true,detect-inline-weight=math}
  \begin{tabular}{lccS[table-format=3.1]S[table-format=3.1]}
  \toprule
  \multirow{2}{*}{\bfseries Method} & \multirow{2}{*}{\bfseries R} & \multirow{2}{*}{\bfseries p} & \bfseries {MPJPE [mm]} & \bfseries {MPJPE [mm]}\\
  & & & \bfseries {(UDA)} & \bfseries {(SFDA)}\\
  \midrule
  no filtering & - & - & 102.3 & 100.8\\
  consistency & - & - & 102.1 & 100.5\\
  \midrule
  $\mathcal{L}_{\mathrm{angle}}$ & 0.20 & $<10^{-3}$ & 100.1 & 102.8\\
  $\mathcal{L}_{\mathrm{sym}}$ & 0.36 & $<10^{-3}$ & 99.6 & 98.4\\
  $\mathcal{L}_{\mathrm{\textcolor{rev_changes}{length}}}$ & 0.56 & $<10^{-3}$ & 92.9 & 98.7\\
  $\mathcal{L}_{\mathrm{anat}}$ & 0.45 & $<10^{-3}$ & 93.1 & 97.9\\
  2 out of 3 & - & - & \bfseries 92.3 & \bfseries 97.0\\
  \bottomrule
  \end{tabular}
\end{table}
To this end, we use the source-only model for inference on the validation set of the target domain.
\begin{table*}[t]
  \caption{\label{Tab:SOTA-SLP}Results for adaptation in the uncover$\rightarrow$cover setting on the SLP dataset for both UDA and SFDA methods. We compare the MPJPE [mm] of our method to diverse competing methods. Results are averaged over thin and thick cover as the scores are almost identical. Mean\textsuperscript{*} indicates the average over the joints shared with the MVIBP dataset, namely feet, knees, shoulders, elbows, and hands.}
  \centering
  \sisetup{detect-weight=true,detect-inline-weight=math}
  \begin{tabular}{lccS[table-format=3.1]S[table-format=3.1]S[table-format=2.1]S[table-format=2.1]S[table-format=3.1]S[table-format=3.1]S[table-format=3.1]S[table-format=3.1]S[table-format=3.1]S[table-format=3.1]}
  \toprule
  \bfseries Method & \bfseries UDA & \bfseries SFDA & \bfseries {Feet} & \bfseries {Knees} & \bfseries {Hips} & \bfseries {Core} & \bfseries {Head} & \bfseries {Shoul} & \bfseries {Elb} & \bfseries {Hands} & \bfseries {Mean\textsuperscript{*}} & \bfseries {Mean}\\
  \midrule
  mean pose & & & 239.4 & 240.0 & 56.1 & 31.8 & 102.7 & 134.6 & 292.6 & 383.0 & 257.9 & 189.1\\
  source-only & & & 174.1 & 148.1 & 74.5 & 56.5 & 34.8 & 65.7 & 168.2 & 273.2 & 165.9 & 130.4\\
  target-only & & & 86.4 & 64.8 & 36.7 & 31.6 & 29.4 & 42.3 & 80.6 & 140.0 & 82.8 & 67.7\\
  \midrule
  MMD & \checkmark & & 164.6 & 124.6 & 68.5 & 56.9 & 35.3 & 62.8 & 177.1 & 243.0 & 154.4 & 121.7\\
  DANN & \checkmark & & 168.8 & 114.5 & 60.9 & 50.3 & 33.3 & 55.0 & 144.8 & 218.8 & 140.4 & 111.6\\
  DefRec & \checkmark & & 161.0 & 130.6 & 68.1 & 51.4 & 34.5 & 63.6 & 175.3 & 255.0 & 157.1 & 122.6\\
  SSDispPred & \checkmark & & 168.4 & 122.7 & 65.7 & 51.0 & 33.9 & 59.9 & 165.1 & 258.4 & 154.9 & 121.9\\
  AdvOutAdapt & \checkmark & & 181.4 & 128.6 & 62.9 & \bfseries 47.1 & 35.5 & 59.3 & 136.8 & 207.9 & 142.8 & 112.9\\
  CC-SSL & \checkmark & & 144.9 & 134.1 & 71.7 & 54.9 & 33.6 & 59.7 & 145.4 & 222.3 & 141.3 & 112.4\\
  MCD & \checkmark & & 151.8 & 116.8 & 63.7 & 52.6 & 33.6 & 53.1 & 120.4 & 171.4 & 122.7 & 99.4\\
  Mean Teacher & \checkmark & & 155.9 & 109.8 & 73.6 & 57.4 & 35.0 & 56.1 & 118.6 & 175.9 & 123.3 & 102.3\\
  \midrule
  ours, $\mathcal{L}_{\mathrm{anat}}$ only & \checkmark & & 141.5 & 102.2 & \bfseries 56.0 & 47.2 & 33.3 & \bfseries 50.4 & 112.5 & 188.4 & 119.0 & 96.6\\
  ours, $\mathcal{L}_{\mathrm{con}}$ only & \checkmark & & 134.7 & 97.3 & 60.0 & 49.1 & \bfseries 33.2 & 54.3 & 110.5 & \bfseries 163.9 & 112.1 & 92.1\\
  ours & \checkmark & & \bfseries 120.4 & \bfseries 97.0 & 57.4 & \bfseries 47.1 & 33.8 & 51.7 & \bfseries 109.1 & 169.8 & \bfseries 109.6 & \bfseries 89.6\\
  \midrule
  UBNA & & \checkmark & 172.9 & 136.1 & 71.3 & 57.2 & 37.3 & 60.4 & 149.1 & 259.8 & 155.7 & 124.5\\
  BNAdapt & & \checkmark & 167.5 & 125.0 & 69.4 & 59.0 & 35.1 & 63.5 & 154.0 & 229.5 & 147.9 & 118.5\\
  Mean Teacher & & \checkmark & 137.4 & 110.6 & 66.2 & 51.6 & \bfseries 32.9 & 56.8 & 134.7 & 186.8 & 125.3 & 100.8\\
  \midrule
  ours, $\mathcal{L}_{\mathrm{anat}}$ only & & \checkmark & 155.6 & 120.0 & 64.2 & 54.6 & 37.0 & 55.7 & 126.0 & 206.7 & 132.8 & 107.7\\
  ours, $\mathcal{L}_{\mathrm{con}}$ only & & \checkmark & 133.1 & 102.8 & 65.1 & 50.9 & 33.2 & 55.5 & 127.4 & \bfseries 178.1 & 119.4 & 97.0\\
  ours & & \checkmark & \bfseries 132.8 & \bfseries 102.5 & \bfseries 62.3 & \bfseries 49.8 & 33.6 & \bfseries 53.1 & \bfseries 118.3 & 179.6 & \bfseries 117.3 & \bfseries 95.6\\
  \bottomrule
  \end{tabular}
\end{table*}
For each predicted pose, we compute the pose error (MPJPE) and the anatomical losses $\mathcal{L}_{\mathrm{sym}},\mathcal{L}_{\mathrm{angle}},\mathcal{L}_{\mathrm{\textcolor{rev_changes}{length}}}$, and $\mathcal{L}_{\mathrm{anat}}$.
We then compute the Pearson coefficient $R$ and the corresponding p-value between pose errors and each of the losses (see Tab.~\ref{Tab:ablation_mean-teacher}, columns 2,3).
For all loss functions, p-values smaller than 0.001 prove a significant correlation, confirming our hypothesis.
Comparing the Pearson coefficients among the individual loss functions, we obtain---from weakest to strongest correlation---$\mathcal{L}_{\mathrm{angle}},\mathcal{L}_{\mathrm{sym}}$, and $\mathcal{L}_{\mathrm{\textcolor{rev_changes}{length}}}$.
Interestingly, this order is identical to model performance when using the loss functions for direct supervision in anatomy-constrained optimization (see Tab.~\ref{Tab:ablation_anatomical}).
The Pearson coefficient for $\mathcal{L}_{\mathrm{anat}}$ ranges between those for $\mathcal{L}_{\mathrm{sym}}$ and $\mathcal{L}_{\mathrm{\textcolor{rev_changes}{length}}}$.

Given the confirmation of our initial hypothesis, we now explore the suitability of the loss functions for filtering pseudo labels.
To this end, we consider both UDA and SFDA settings, discard anatomy-constrained optimization ($\lambda_1=0$), and use different variants of \textcolor{rev_changes}{Eq.~\eqref{Eq:filter-criterion}} to guide the consistency training.
Besides our proposed method (denoted as `2 out of 3'), we filter pseudo labels by directly comparing each of the losses $\mathcal{L}_{\mathrm{sym}},\mathcal{L}_{\mathrm{angle}},\mathcal{L}_{\mathrm{\textcolor{rev_changes}{length}}}$, and $\mathcal{L}_{\mathrm{anat}}$.
As the baseline, we perform no filtering ($h(\hat{\boldsymbol{Y}}',\hat{\boldsymbol{Y}})=1$), which is equivalent to the standard Mean Teacher.
As another comparison method, similar to \citet{ke2019dual,mu2020learning,zhou2022uncertainty}, we filter pseudo labels based on their consistency under input augmentations.
Specifically, we forward two augmented versions of the input through both the student and the teacher model and compute a consistency loss between the two student predictions and the two teacher predictions.
On this basis, the teacher predictions are only used for supervision if they are more consistent than the student predictions.

Results of the experiment are shown in Tab~\ref{Tab:ablation_mean-teacher}, columns 4 and 5.
Our insights are four-fold.
First, consistency-based filtering yields only a minor improvement compared to the baseline without filtering.
Second, as intuitively expected, we observe a rough trend that a higher correlation between the anatomical loss functions and the pose error comes along with improved performance when using the losses for filtering pseudo labels.
Specifically, filtering based on $\mathcal{L}_{\mathrm{angle}}$ yields a minor improvement for UDA and even a slight degradation for SFDA.
Moderate improvements under both scenarios are realized by $\mathcal{L}_{\mathrm{sym}}$, while $\mathcal{L}_{\mathrm{\textcolor{rev_changes}{length}}}$ and $\mathcal{L}_{\mathrm{anat}}$ achieve the top performance among the loss functions.
Third, our proposed `2 out of 3' method further improves on $\mathcal{L}_{\mathrm{\textcolor{rev_changes}{length}}}$ and $\mathcal{L}_{\mathrm{anat}}$.
This indicates that our proposed ensembling strategy of the three individual losses is superior to simple aggregation in $\mathcal{L}_{\mathrm{anat}}$, where different scales of the losses are neglected.
Fourth, our method surpasses the baseline method (no filtering) by 10\% for UDA and by 4\% for SFDA.
Thus, our anatomy-based filtering strategy considerably improves the efficiency of self-training with pseudo labels under the Mean Teacher paradigm.

\subsection{Comparison to the state of the art}
\begin{figure*}[t]
\begin{center}
  \includegraphics[width=\linewidth]{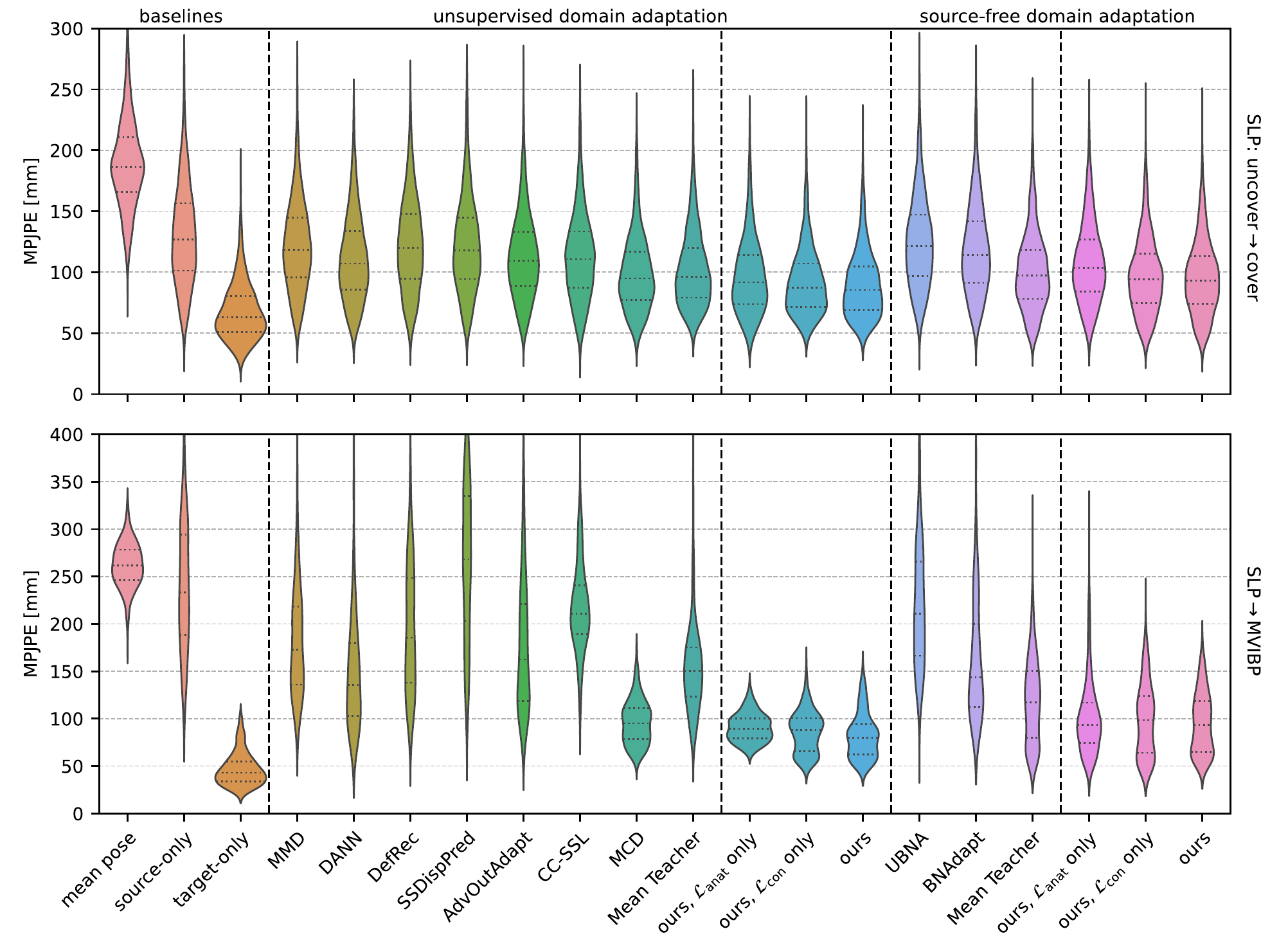}
\end{center}
  \textcolor{rev_changes}{\caption{Violin plots of the frame-averaged joint errors for all compared methods in the uncover$\rightarrow$cover setting on the SLP dataset (top) and for SLP$\rightarrow$MVIBP adaptation (bottom). Dashed lines inside the violins represent the 25th, 50th, and 75th percentiles.}}
\label{Fig:violin_plots}
\end{figure*}

We compare our method to the comparison methods presented in Sec.~\ref{Sec:comparison_methods} under the two adaptation scenarios uncover$\rightarrow$cover (U$\rightarrow$C) and SLP$\rightarrow$MVIBP.
Quantitative results are shown in Tab.~\ref{Tab:SOTA-SLP}, Tab.~\ref{Tab:SOTA-SLP-MVIBP}, \textcolor{rev_changes}{and Fig.~\ref{Fig:violin_plots}}, revealing mostly consistent findings.

First, we note that the mean pose baseline yields an insufficient accuracy under both scenarios, with a similar mean error when averaged over the same set of joints.
\begin{table*}[t]
  \caption{\label{Tab:SOTA-SLP-MVIBP}Results for SLP$\rightarrow$MVIBP adaptation for both UDA and SFDA. We compare the MPJPE [mm] of our method to diverse competing methods.}
  \centering
  \sisetup{detect-weight=true,detect-inline-weight=math}
  \begin{tabular}{lccS[table-format=3.1]S[table-format=3.1]S[table-format=3.1]S[table-format=3.1]S[table-format=3.1]S[table-format=3.1]}
  \toprule
  \bfseries Method & \bfseries UDA & \bfseries SFDA & \bfseries {Feet} & \bfseries {Knees} & \bfseries {Shoul} & \bfseries {Elb} & \bfseries {Hands} & \bfseries {Mean}\\
  \midrule
  mean pose & & & 272.4 & 228.8 & 128.3 & 229.4 & 449.8 & 261.7\\
  source-only & & & 117.2 & 104.5 & 114.0 & 347.0 & 517.3 & 240.0\\
  target-only & & & 55.8 & 37.3 & 32.0 & 36.0 & 73.1 & 46.8\\
  \midrule
  MMD & \checkmark & & 132.4 & 120.0 & 130.1 & 265.0 & 273.6 & 184.2\\
  DANN & \checkmark & & 111.2 & 103.4 & 118.6 & 206.6 & 206.4 & 149.2\\
  DefRec & \checkmark & & 112.6 & 82.8 & 112.1 & 370.9 & 296.3 & 194.9\\
  SSDispPred & \checkmark & & 148.1 & 99.8 & 132.4 & 355.2 & 600.7 & 267.2\\
  AdvOutAdapt & \checkmark & & 157.3 & 141.6 & 101.4 & 189.7 & 297.3 & 177.4\\
  CC-SSL & \checkmark & & 85.7 & 73.4 & 90.1 & 321.1 & 515.7 & 217.2\\
  MCD & \checkmark & & 89.6 & 77.1 & \bfseries 61.5 & 92.2 & 161.9 & 96.4\\
  Mean Teacher & \checkmark & & 93.4 & 77.3 & 110.6 & 291.6 & 197.0 & 154.0\\
  \midrule
  ours, $\mathcal{L}_{\mathrm{anat}}$ only & \checkmark & & 86.6 & 85.2 & 70.7 & 85.6 & 126.0 & 90.8\\
  ours, $\mathcal{L}_{\mathrm{con}}$ only & \checkmark & & 63.0 & 70.5 & 117.9 & \bfseries 75.6 & \bfseries 103.0 & 86.0\\
  ours & \checkmark & & \bfseries 62.6 & \bfseries 70.2 & 83.2 & 84.9 & 108.4 & \bfseries 81.8\\
  \midrule
  UBNA & & \checkmark & 101.0 & 102.5 & 130.2 & 371.3 & 386.1 & 218.2\\
  BNAdapt & & \checkmark & 96.3 & 108.5 & 142.2 & 223.7 & 231.1 & 160.4\\
  Mean Teacher & & \checkmark & 84.4 & 72.2 & 109.1 & 131.8 & 194.6 & 118.4\\
  \midrule
  ours, $\mathcal{L}_{\mathrm{anat}}$ only & & \checkmark & 81.7 & 88.8 & \bfseries 75.3 & 106.2 & \bfseries 156.8 & 101.8\\
  ours, $\mathcal{L}_{\mathrm{con}}$ only & & \checkmark & \bfseries 66.7 & \bfseries 62.5 & 83.7 & 98.4 & 182.9 & 98.8\\
  ours & & \checkmark & 68.8 & 68.3 & 79.3 & \bfseries 86.8 & 174.4 & \bfseries 95.5\\
  \bottomrule
  \end{tabular}
\end{table*}
This indicates a comparable difficulty and variability of poses across the SLP and MVIBP datasets.
Note that the low error for hip and core joints for U$\rightarrow$C is due to their spatial proximity to the root joint whose ground truth position was used at inference time.

\begin{figure*}[t]
\begin{center}
  \includegraphics[width=\linewidth]{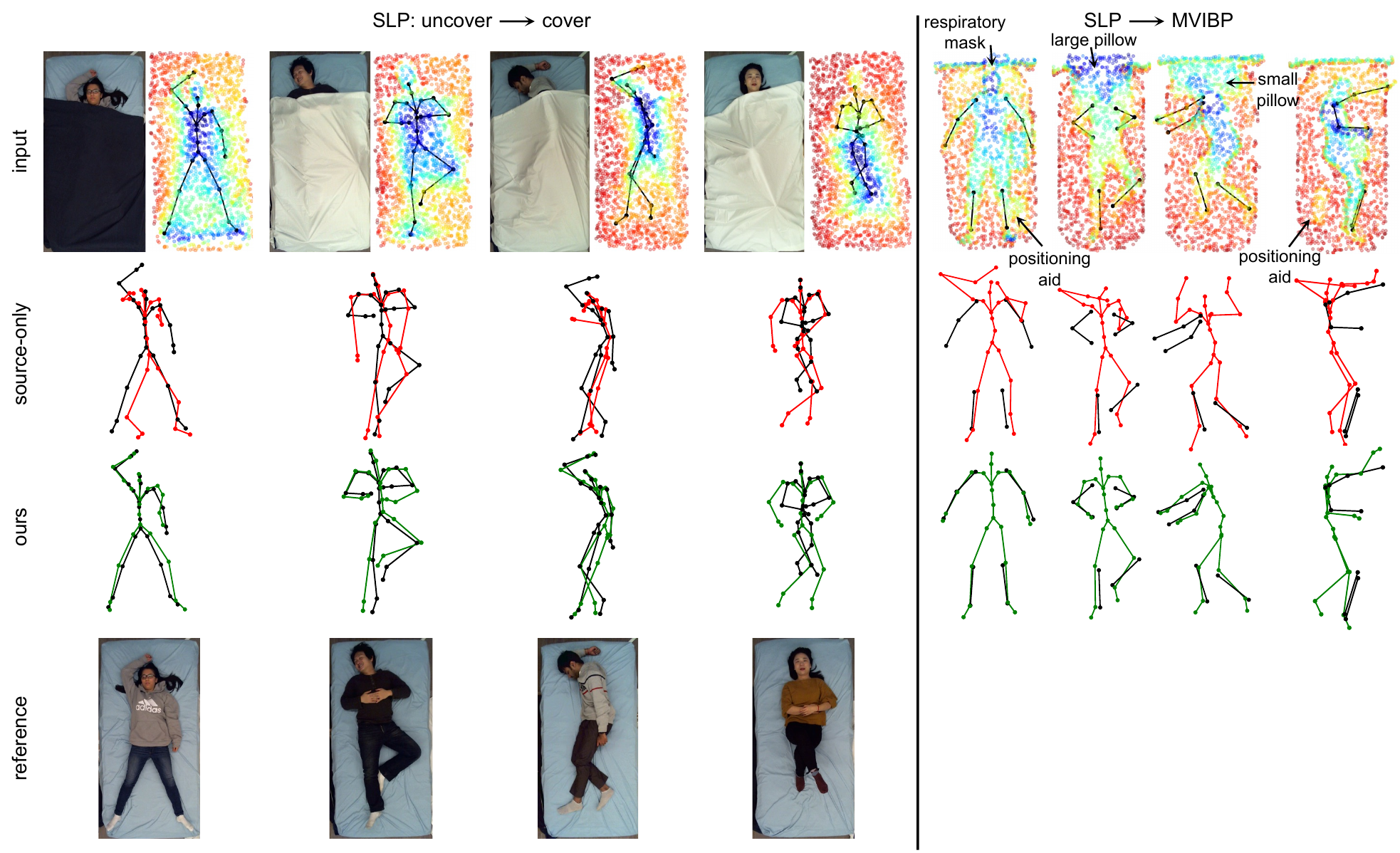}
\end{center}
  \caption{Qualitative results on test samples from the target domain for uncover$\rightarrow$cover adaptation (columns 1-4) and for SLP$\rightarrow$MVIBP adaptation (columns 5-8).
  We show input point clouds (upper row), predictions by the source-only model (second row, red), and predictions by our method (third row, green) each together with the ground truth pose in black.
  For the samples from the SLP dataset, we also show the color images belonging to the point clouds (first row) for better visualization and the corresponding color images without a cover (fourth row) for reference.
  Regarding the MVIBP dataset, we must not show any color images due to data privacy.}
\label{Fig:qualitative_results}
\end{figure*}

Second, the source-only baseline is far superior to the mean pose estimate but still substantially worse than the target-only oracle.
Specifically, the MPJPE of the target-only model is increased by 93\% for  U$\rightarrow$C (100\% when averaged over the joints shared with MVIBP) and by even 413\% for SLP$\rightarrow$MVIBP.
This confirms that both considered domain shifts pose severe problems for deep learning-based pose estimation models.
Interestingly, the domain shift due to the occlusion by a cover, which intuitively appears more severe to humans than the shift between the two datasets, has a substantially less negative impact on model performance.
We also observe that the MPJPE for shoulders, elbows, and hands of the source-only model is higher for SLP$\rightarrow$MVIBP than for U$\rightarrow$C.
The reason presumably is that the domain shift for SLP$\rightarrow$MVIBP is partially caused by the presence of a headboard and pillows, which mainly complicate the localization of joints in the upper body (see Fig.~\ref{Fig:qualitative_results}, rows 5-7).
Meanwhile, the MPJPE for feet and knees of the source-only model is lower for SLP$\rightarrow$MVIBP, and the oracle achieves lower errors for all joints for SLP$\rightarrow$MVIBP.
These two observations, in turn, are likely due to the absence of a blanket in this scenario, simplifying the pose estimation problem, especially for joints of the lower body.

Third, we assess the performance of the state-of-the-art comparison methods and our method for UDA.
All comparison methods improve the source-only model under both scenarios, except for SSDispPred, which fails for SLP$\rightarrow$MVIBP.
The ranking of the methods is also similar under both domain shifts (only CC-SSL is less effective for SLP$\rightarrow$MVIBP), with MCD achieving the lowest error.
Most importantly, the results show that both of our proposed methods alone, i.e., anatomy-constrained optimization ($\mathcal{L}_{\mathrm{anat}}$ only) and anatomy-guided filtering of pseudo labels ($\mathcal{L}_{\mathrm{con}}$ only), already outperform all comparison methods under both settings, with $\mathcal{L}_{\mathrm{con}}$ only being slightly superior to $\mathcal{L}_{\mathrm{anat}}$ only.
Notably, our anatomy-constrained optimization surpasses adversarial output adaptation, highlighting the effectiveness of explicit constraints contrary to adversarial optimization.
The results further show that our two methods are complementary as their combination further reduces the MPJPE to \SI{89.6}{mm} for  U$\rightarrow$C and \SI{81.8}{mm} for SLP$\rightarrow$MVIBP.
This corresponds to a relative improvement of 31\% and 66\% over the source-only model and a reduction of the gap between the source-only and the target-only model of 65\% and 82\%, respectively.

Finally, we compare the SFDA methods.
Again, each comparison method reduces the domain gap under both settings.
Among these methods, the Mean Teacher achieves the highest performance, surprisingly outperforming its counterpart for UDA.
As possible reasons, we suspect the frozen weights of the network heads and a better adaptation of the BatchNorm statistics in SFDA.
Regarding our proposed methods, we make the expected observation that all three versions perform slightly worse than in the UDA setting.
Nevertheless, they are still superior to the comparison methods for SFDA (except $\mathcal{L}_{\mathrm{anat}}$ only, which is inferior to the Mean Teacher for U$\rightarrow$C), and---importantly---the combined method is even superior to all competing UDA methods under both domain shifts.
This demonstrates the high efficiency of our method under the challenging SFDA setting.

Qualitative results are shown in Fig.~\ref{Fig:qualitative_results} and are consistent with the quantitative findings.
Both the occlusion by a blanket (columns 1-4) and the presence of medical/bed utils (positioning aid, pillow, respiratory mask; columns 5-8) confuse the source-only model, which predicts inaccurate and anatomically implausible poses.
By contrast, the predictions by our anatomy-guided adaptation method are more accurate and anatomically more plausible.
In particular, our method prevents implausible bone lengths in arms (columns 1,2,3,5,6,7) and legs (columns 1,2) and implausible angles in the shoulder, elbow, and wrist joints (columns 1,3,5,6).
Two failure cases of our method are shown in columns 4 and 8, where the predicted poses appear plausible but are inconsistent with the actual pose.

\section{Discussion and conclusion}
We introduced a novel domain adaptation method for point cloud-based 3D human pose estimation.
Our main methodological contribution is to bridge the domain gap with the aid of prior anatomical knowledge, accomplished by two complementary anatomy-based adaptation strategies. 
First, we directly supervise target predictions by imposing explicit anatomical constraints on the output space.
Second, we filter pseudo labels for self-training according to their anatomical plausibility.
Our experiments for in-bed pose estimation confirm the efficacy of both approaches and allow the following conclusions:
1) Anatomical constraints are a powerful source of weak supervision to guide the learning process in the absence of ground truth.
2) Anatomy-based filtering of pseudo labels substantially improves the efficiency of self-training.
Specifically, we evaluated our method under two different domain shifts, adapting from uncovered to covered subjects and between the different environments of two datasets.
In both settings, our method outperformed diverse comparison methods, surpassed the baseline model by 31\%/66\%, and reduced the domain gap by 65\%/82\%.
In absolute terms, it reduced the mean error of pose estimates to less than \SI{9}{cm} for covered patients and to almost \SI{8}{cm} for uncovered patients.
At the same time, our method proved efficient for both UDA and SFDA, thus enabling adaptation even in case of restricted data access.
In summary, our method can avoid the need for costly manual annotations in novel target domains, which is a significant obstacle to the flexible use of pose estimation models.
Thus, it could become an essential factor in advancing the practical deployment of clinical monitoring systems.

Considering this intended application in a realistic clinical setting, a more detailed discussion of the outcomes of our study is needed.
First, while the reported results by our method for pose estimation are promising, in practice, we are interested in the performance of higher-level downstream tasks like action or posture recognition.
In consequence, the following open questions still need to be analyzed in future clinical validation studies:
To what extent do the improvements by our method enhance the performance of different downstream tasks?
Is the pose accuracy by our method sufficient, or are there any downstream tasks that require higher accuracy?
Second, the evaluation in our work is restricted to healthy subjects in both domains.
However, when adapting a model from a lab dataset (source) to clinical data (target), we might face a population shift, with clinical patients showing pathologically induced anatomical abnormalities, such as asymmetric or deformed limbs.
Our symmetry and bone length losses in their original form ($\delta_i=0$, bounds derived from the source data) would then provide incorrect supervision and no longer be a suitable criterion for filtering pseudo labels.
This, in turn, might hamper the general adaptation process and degrade pose estimates for pathological patients.
A similar problem would occur when adapting from adults in the source to children (at the pediatric ward, for instance) in the target domain.
Advantageously, the formulation of our method is flexible enough to prevent such problems by carefully adjusting the upper and lower bounds of symmetry and bone length constraints according to the target population.
The bone lengths of children could be looked up in an anatomical textbook, and patient-specific bounds would enable the incorporation of patient-specific anatomical abnormalities.

As a methodological outlook, we see multiple further opportunities for the beneficial use of anatomical priors.
First, our anatomical constraints only approximate the space of plausible poses, still permitting implausible poses.
This is mainly caused by our realization of the constraint on the joint angles: 1) Joints are considered in isolation, ignoring the pose dependency of joint limits \citep{akhter2015pose}, and 2) the used scalar product does not uniquely represent 3D angles.
Incorporating a kinematic model could alleviate these problems and help enforce a globally plausible joint-angle configuration.
Second, imposing anatomical constraints during training does not preclude implausible pose estimates at inference time.
Embedding the constraints in the model architecture itself, instead, would eliminate implausible predictions and could thus increase model robustness.
Third, in the context of patient monitoring, we have access to a continuous stream of input data instead of isolated frames, opening up further options for using anatomical priors.
On the one hand, we can exploit confident pose estimates to derive approximate patient-specific bone lengths.
These could serve as prior knowledge to guide the pose estimation on subsequent frames of the same patients, for instance, by conditioning the model on their specific anatomy.
On the other hand, a model that operates on a sequence of successive frames could be constrained to predict anatomically coherent poses across time.

Finally, beyond the specific task of point cloud-based human pose estimation, our method might also be beneficial for domain adaptation in general medical imaging tasks.
On the one hand, anatomy-constrained optimization could be adapted to 3D landmark detection tasks.
On the other hand, the filtering of pseudo labels according to explicit prior knowledge about the structure of the output space is---to the best of our knowledge---a novel concept transferable to other tasks.
Pseudo labels in medical segmentation, for instance, could be filtered according to prior knowledge about shape descriptors \citep{bateson2022test,kervadec2021beyond}.

\section*{Acknowledgments}
We gratefully acknowledge the financial support by the Federal Ministry for Economic Affairs and Climate Action of Germany (FKZ: 01MK20012B).

\bibliographystyle{model2-names.bst}\biboptions{authoryear}
\bibliography{refs}

\end{document}